# A robust method for reliability updating with equality information using sequential adaptive importance sampling

Xiong Xiao[1], Zeyu Wang[1], Quanwang Li[1*]

[1] Department of Civil Engineering, Tsinghua University, Beijing 100084, China

## Abstract

Reliability updating refers to a problem that integrates Bayesian updating technique with structural reliability analysis and cannot be directly solved by structural reliability methods (SRMs) when it involves equality information. The state-of-the-art approaches transform equality information into inequality information by introducing an auxiliary standard normal parameter. These methods, however, encounter the loss of computational efficiency due to the difficulty in finding the maximum of the likelihood function, the large coefficient of variation (COV) associated with the posterior failure probability and the inapplicability to dynamic updating problems where new information is constantly available. To overcome these limitations, this paper proposes an innovative method called RU-SAIS (reliability updating using sequential adaptive importance sampling), which combines elements of sequential importance sampling and $K$-means clustering to construct a series of important sampling densities (ISDs) using Gaussian mixture. The last ISD of the sequence is further adaptively modified through application of the cross entropy method. The performance of RU-SAIS is demonstrated by three examples. Results show that RU-SAIS achieves a more accurate and robust estimator of the posterior failure probability than the existing methods such as subset simulation.

## 1. Introduction

Structural reliability has been used as a probability measure of structural safety since the middle of last century, which can be defined as $1- p_f$, where $p_f$ is the probability of structure failure when the external load $S$ exceeds the structural resistance $R$. Both of them are modeled as probabilistic variables parameterized by a set of random variables $\boldsymbol{X} = [X_1; X_2; …; X_n]$ with the joint probability density function (PDF) $\pi(\boldsymbol{x})$ (In this paper, a multidimensional variable is represented by a bold uppercase letter, and its realization is represented by the corresponding bold lowercase letter). Therefore, the failure probability $p_f$ is calculated by

$$p_{\text{f}} = \Pr(E) = \int_{x \in \Omega_E} \pi(x)\,\mathrm{d}x = \int_x I[g(x) \leq 0]\pi(x)\,\mathrm{d}x, \tag{1}$$

where $E$ and $\Omega_E$ respectively denote the failure event and its corresponding domain in the outcome space of $X$, $g(x)$ denotes the limit state function written as $g(x) = R(x) - S(x)$, $I(\cdot)$ denotes the indicator function that takes one if $g(x) \leq 0$ and 0 otherwise throughout the whole space $\mathbb{R}^n$, and $\int_x \mathrm{d}x$ is the short form of $\int_{x \in \mathbb{R}^n} \mathrm{d}x$, which is used throughout this paper. Many structural reliability methods (SRMs) have been developed in the past decades to resolve the complicated $n$-dimensional integrals in Eq. (1) [1]–[6]. However, the resulting failure probability is of uncertainty since $\pi(x)$ is assumed according to empirical judgement in most cases. Reliability updating enables to reduce the uncertainties in calculating $p_{\text{f}}$ when new information or data measured in-site is available. This can be achieved from two different perspectives. One viewpoint is to directly calculate the conditional probability

$$p_{\text{f}}' = \Pr(E \mid Z) = \frac{\Pr(E \cap Z)}{\Pr(Z)} = \frac{\int_{x \in \{\Omega_E \cap \Omega_Z\}} \pi(x)\,\mathrm{d}x}{\int_{x \in \Omega_Z} \pi(x)\,\mathrm{d}x}, \tag{2}$$

where $p_{\text{f}}'$ is the updated (posterior) failure probability (in this context, $p_{\text{f}}$ is called the prior failure probability), $Z = Z_1 \cap Z_2 \cap \cdots \cap Z_m$ and $\Omega_Z = \Omega_{Z_1} \cap \Omega_{Z_2} \cap \ldots \cap \Omega_{Z_m}$ respectively denote the event of $m$ observations and its corresponding domain in the outcome space of $X$. The observation $Z_i$ is of inequality type when expressed as $\Omega_{Z_i} = \{h_i(x) \leq 0\}$ and equality type when expressed as $\Omega_{Z_i} = \{h_i(x) = 0\}$, where $h_i(x)$ is the limit state function associated with it [7]. When all of the $m$ observations are of the inequality type, SRMs suitable for solving Eq. (1) can be used to solve the numerator and denominator of Eq. (2), respectively, and then the posterior failure probability is obtained. However, in the case of one or more observations being of the equality type, both the numerator and denominator result 0, which makes Eq. (2) invalid.

Several solutions have been appeared in literature to resolve the above problem. One strategy is replacing the volume integrals ($n \geq 3$) with surface integrals. In program STRUREL, the traditional FORM and SORM (first- and second- order reliability method) are integrated with surface integration techniques to estimate Eq. (2) with equality observation(s) [8]. This solution is fast but inherits the defects of FORM and SORM that the accuracy deteriorates dramatically for highly nonlinear problems. Another strategy is to transform this problem so that the volume integrals are still available. By introducing a dummy variable $\delta$, the posterior reliability with equality information $\{h(x) = 0\}$ can be expressed as: [7], [9]

$$\Pr(g(\bm{x}) \le 0 \mid h(\bm{x}) = 0) = \frac{\lim\limits_{\delta \to 0} \dfrac{\partial}{\partial \delta} \Pr(g(\bm{x}) \le 0 \cap h(\bm{x}) - \delta \le 0)}{\lim\limits_{\delta \to 0} \dfrac{\partial}{\partial \delta} \Pr(h(\bm{x}) - \delta \le 0)}, \tag{3}$$

which enables the using of SRMs. The method is feasible in theory but can cause significant error because most SRMs cannot meet the accuracy requirement of calculating the partial derivative numerically. Alternatively, Straub rewrote the equality information as inequality information and subtly turned the problem into two general reliability analysis problems [7] by splitting *X* into two variables and introducing an auxiliary standard normal variable *U*. This approach combined with subset simulation [4] has been successfully used in deteriorating systems [10] and geotechnical engineering [11]. However, some limitations affect the computation performance of this method. The inherit shortcomings of Markov chain Monte Carlo (MCMC) such as sample correlation and burn-in period lead to large coefficient of variation (COV). Furthermore, an appropriate selection for the multiplier *c* is required, which defines the observation domain and is usually difficult when the number of observations is large. A conservative value of *c* will make the failure probabilities of both the numerator and denominator extremely small and thus increases the number of intermediate subsets. Meanwhile, *c* should be re-chosen and these two failure probabilities should be re-evaluated from scratch when new information is available. Improvements to some of these limitations have been appeared in [12]–[14]. In contrast to subset simulation, Wang and Shafieezadeh proposed the RUAK (reliability updating with adaptive Kriging) method to improve the computation performance of Straub's method [15], which significantly reduces computational cost by replacing the time-consuming instrumented model with a trained Kriging model. However, their method only constrains the COV of the prior failure probability while the COV of the posterior one is unknown. Moreover, Kriging-based methods [16] have some inherit disadvantages; on the one hand, they face challenges in dealing with problems with very rare failure event or highly nonlinear failure surface; on the other, it is difficult for non-experts to obtain a well-trained Kriging model since the training process requires empirical judgement.

In contrast to the above viewpoint that directly calculating $p_\text{f}'$, the other perspective is updating the joint PDF of *X* firstly, and then utilizing the posterior density *p*(*x*) to estimate $p_\text{f}'$, i.e.,

$$p(\bm{x}) = \Pr(\bm{X} = \bm{x} \mid Z),\ p_\text{f}' = \int_x I[g(\bm{x}) \le 0] p(\bm{x}) \mathrm{d}\bm{x}. \tag{4}$$

It has been proven that the perspective is theoretically equivalent to the previous one [15]. Starting with this viewpoint, several methods have been proposed in literature [17]–[19]. The integral in Eq. (4) is calculated in [20] by replacing *p*(*x*) with an asymptotic approximation, which is suitable only when $p(\boldsymbol{x})$ is sharply peaked at isolated points. Beck and Au updated the failure probability using Markov chain samples approximating $p(\boldsymbol{x})$ [21]. These samples were generated with an adaptive Metropolis-Hastings (AMH) algorithm, which motivated Ching and Chen to further develop the transitional Markov chain Monte Carlo (TMCMC) method [22]. The computation performance of both methods get worse as the $p'_f$ decreases. Further along this strategy, the Markov chain samples were adopted to generate a new PDF $q(\boldsymbol{x})$ and the corresponding cumulative distribution function (CDF) $Q(\boldsymbol{x})$. This is followed by the standard FORM to arrive at the estimation of $p'_f$ [23]. The method, involving sampling, fitting and approximation, inevitably leads to large errors when $p(\boldsymbol{x})$ or $g(\boldsymbol{x})$ is complex.

To overcome the foregoing limitations, this paper develops a novel sequential adaptive importance sampling algorithm for robust (i.e., the resulting estimator is accurate and has small COV) reliability updating with equality information. This method, called RU-SAIS, constructs a sequence of Gaussian mixture densities by integrating the elements of sequential importance sampling and $K$-means clustering algorithm. In this context, a series of intermediate densities are determined step by step with the prior density as the first one and the optimal important sampling density (ISD) as the last one. The corresponding sequence of Gaussian mixture densities are constructed by using $K$-means clustering. In each step, the Gaussian mixture density is a rough approximation to the intermediate density and has "heavier" tails. When the intermediate density achieves or is very close to the optimal one, the sequential adaptive importance sampling procedure is terminated and the last Gaussian mixture density is then further modified with cross entropy method. The estimated COVs of the estimators of the numerator and denominator indicated in Eq. (2) are selected as the stopping criterion of RU-SAIS. Three examples are investigated to demonstrate the computation performance of the proposed method. These include a two-dimensional numerical example, a ten-dimensional truss system and an engineering-guided chloride and carbonation-induced concrete corrosion case.

This paper starts with the key idea of importance sampling-based reliability updating in section 2, followed by the detailed steps of the proposed RU-SAIS method and its flowchart in section 3. The statistical properties of the estimators are derived in section 4 and three illustrative examples are presented in section 5 to investigate the computation performance of the proposed method.

## 2. Reliability updating using importance sampling

This paper proposes to conduct reliability updating with importance sampling technique, and the proposed method is based on the second viewpoint introduced in section 1. It retains the strategy in Straub's method [7] that partitioning $X$ into two variables, i.e., $X$

= [$X_g$; $X_h$], where $X_h$ denotes the variable(s) only presented in the response function $h(x)$ and $X_g$ denotes the other ones in $X$. $X_h$ is typically a scalar and refers to the measurement error $\varepsilon$, which is often assumed to be Gaussian distributed with mean 0. In this context, the equality information can be expressed as $\Omega_Z = \{h(x_g, \varepsilon) = s(x_g) - s_m + \varepsilon = 0\}$ and the likelihood function representing its effect on reliability updating is

$$L(x_g) \propto \Pr(Z | X_g = x_g) \propto f_\varepsilon \left[ s(x_g) - s_m \right], \tag{5}$$

where $s(x_g)$ and $s_m$ respectively denote the response function and its corresponding measurement, and $f_\varepsilon$ denotes the PDF of $\varepsilon$. According to Bayes theorem, the posterior density is derived as

$$p(x_g) = \Pr(X_g = x_g / Z) = \frac{\Pr(Z | X_g = x_g) \pi(x_g)}{\Pr(Z)} = \frac{L(x_g) \pi(x_g)}{\int_{x_g} L(x_g) \pi(x_g) \, dx_g}. \tag{6}$$

Substituting Eq. (6) into Eq. (4), one can get

$$p'_f = \frac{\int_{x_g} I\left[ g(x_g) \leq 0 \right] L(x_g) \pi(x_g) \, dx_g}{\int_{x_g} L(x_g) \pi(x_g) \, dx_g}. \tag{7}$$

Denote the numerator and denominator in Eq. (7) as $I_1$ and $I_2$, respectively. Throughout this paper, parameters or functions with a subscript of 1 and 2 are related with the numerator and denominator, respectively. Both of $I_1$ and $I_2$ can be estimated using importance sampling, which requires two proposal ISDs, denoted by $p_1(x_g)$ and $p_2(x_g)$, respectively. The resulting estimator of $p'_f$ is thus

$$\hat{p}'_f = \frac{\hat{I}_1}{\hat{I}_2} = \frac{\frac{1}{N_1} \sum_{k_1=1}^{N_1} \frac{I\left[ g(x_g^{(k_1)}) \leq 0 \right] L(x_g^{(k_1)}) \pi(x_g^{(k_1)})}{p_1(x_g^{(k_1)})}}{\frac{1}{N_2} \sum_{k_2=1}^{N_2} \frac{L(x_g^{(k_2)}) \pi(x_g^{(k_2)})}{p_2(x_g^{(k_2)})}}, \tag{8}$$

where $\hat{p}'_f$, $\hat{I}_1$ and $\hat{I}_2$ respectively denote the estimators of $p'_f$, $I_1$ and $I_2$, $N_1$ and $N_2$ respectively denote the numbers of importance samples generated from $p_1(x_g)$ and $p_2(x_g)$, $x_g^{(k_1)}$ and $x_g^{(k_2)}$ respectively denote the $k_1$-th sample generated from $p_1(x_g)$ and the $k_2$-th sample generated from $p_2(x_g)$. The computation performance of the IS technique largely depends on the choice of the ISD. The ISD is optimal when the variance of estimator becomes 0 [24]. For $I_1$ and $I_2$, the optimal ISDs are respectively

$$p_{1,\text{opt}}(x_g) = \frac{I[g(x_g) \leq 0] L(x_g) \pi(x_g)}{\int_{x_g} I[g(x_g) \leq 0] L(x_g) \pi(x_g) dx_g}, \tag{9}$$

$$p_{2,\text{opt}}(x_g) = \frac{L(x_g) \pi(x_g)}{\int_{x_g} L(x_g) \pi(x_g) dx_g} = p(x_g). \tag{10}$$

It is indicated by Eq. (10) that the optimal ISD for $I_2$ is exactly the posterior density in Eq. (6). However, $p_{1,\text{opt}}(x_g)$ and $p_{2,\text{opt}}(x_g)$ cannot be directly used due to the intractable denominators, instead two near-optimal ISDs resembling them are used.

In the scope of structural reliability analysis and Bayesian updating, there exist many schemes to construct an effective near-optimal ISD, most of which are based on an adaptive importance sampling procedure. Au and Beck proposed to construct a kernel density from the samples simulated from MCMC [3]. Kurtz and Song used a Gaussian mixture and updated the parameters based on the cross entropy method [25]. Xiao et al. proposed to construct a Gaussian mixture located at the local maxima of the optimal ISD and the population of important samples [26]. Engel et al. proposed to construct the optimal ISD using a chosen parametric distribution and identify its parameters through an adaptive sampling approach based on the cross entropy method [27]. However, these methods are developed for either a single Bayesian updating problem or a single reliability analysis problem, thus are computationally inefficient in reliability updating problems. For the former situation, the Markov chain samples or importance samples of the posterior density are inefficient to estimate the posterior failure probability; for the latter situation, the adaptive importance sampling algorithms do not involve the sophisticated likelihood function, thus may get large errors or even unusable in reliability updating problems. In this paper, an adaptive importance sampling method called RU-SAIS is tailored for reliability updating. Of course it is also suitable for single reliability analysis or Bayesian updating problems. When it applies to the single Bayesian updating problem such as estimating the denominator of Eq. (7), it shares some similarities with [26] and [27]. The main difference between RU-SAIS and [26] lies in the approaches of arriving at the crude approximation of the optimal ISD. The proposed method does not need to solve the local optimization problem but is not straightforward enough. Whereas, the difference between RU-SAIS and [27] mainly lies in the way the cross entropy method is used. The proposed method has the advantage that it constructs a Gaussian mixture density with "heavier" tails than the intermediate density in each step, while the drawback is that it cannot be directly integrated with arbitrary parametric densities.

# 3. The proposed RU-SAIS method

In this section, the proposed RU-SAIS method is described in details. We first introduce the idea of sequential adaptive importance sampling and the determination method for the sequence of intermediate densities; the construction scheme of Gaussian mixture density approximating the intermediate density is presented next; the cross entropy method is adopted subsequently to reduce the difference between the Gaussian mixture and the optimal ISD; and finally the stopping criteria of this method is discussed. It is noted that the following functions parameterized by $x_g$ are described in the independent standard normal space, i.e., $\pi(x_g) = \varphi_{n_g}(x_g)$, where $\varphi_{n_g}(\cdot)$ is the $n_g$-variate independent standard normal PDF (here suppose that the dimension of $x_g$ is $n_g$). An isoprobabilistic transformation such as Nataf transformation [28] or Rosenblatt transformation [29] can be performed when random parameters do not follow independent standard normal distribution a priori.

*3.1. Sequential adaptive importance sampling*

Sequential importance sampling [30] provides an innovative approach to address the problems that the optimal ISD $p_{\text{opt}}(x_g)$ is significantly different with the original sampling density $\pi(x_g)$. This technique steams from the particle filter method and is typically integrated with MCMC to resolve the sample impoverishment problem [31], [32]. The key step of sequential importance sampling is to choose a sequence of intermediate densities

$$p^{(i)}(x_g) = \frac{\eta^{(i)}(x_g)}{P^{(i)}}, i = 0, 1, ..., M, \tag{11}$$

where $p^{(0)}(x_g) = \pi(x_g)$, $p^{(M)}(x_g) = p_{\text{opt}}(x_g)$ and $M$ is the number of intermediate densities. In each density, $\eta^{(i)}(x_g)$ is known and $P^{(i)} = \int_{x_g} \eta^{(i)}(x_g) dx_g$ is an unknown normalized constant. In this context, $\eta_1^{(0)}(x_g) = \eta_2^{(0)}(x_g) = \pi(x_g)$, $\eta_1^{(M)}(x_g) = I[g(x_g) \leq 0] L(x_g) \pi(x_g)$ and $\eta_2^{(M)}(x_g) = L(x_g) \pi(x_g)$. Given the densities indicated in Eq. (11), sequential importance sampling enables to draw samples from them in a step-wise manner. When a set of samples from $p^{(i-1)}(x_g)$ are available, an importance resampling procedure and a MCMC algorithm are conducted in succession to result in a new set of samples approximately distributed as $p^{(i)}(x_g)$. Traverse this step for each $i$ ($i = 1, 2, ..., M$), and finally one obtains a set of samples approximately distributed as $p_{\text{opt}}(x_g)$. These samples are correlated since they are drawn from several Markov chains. Moreover, each intermediate step increases the error of the resulting estimator.

The elements of sequential importance sampling described above are adopted in this paper to adaptively construct a sequence of Gaussian mixture densities, which are used as the ISDs in the intermediate steps. This method is thus termed sequential adaptive importance sampling (SAIS). In step $i$ ($i$ = 1, 2, …, $M$), a set of samples from the previous Gaussian mixture density $p_G^{(i-1)}(\mathbf{x}_g)$ (here $\pi(\mathbf{x}_g)$ is regarded as $p_G^{(0)}(\mathbf{x}_g)$) are weighted by $p^{(i)}(\mathbf{x}_g)/p_G^{(i-1)}(\mathbf{x}_g)$ and utilized to construct the current Gaussian mixture density $p_G^{(i)}(\mathbf{x}_g)$. The construction scheme will be introduced in section 3.2. To ensure that $p_G^{(i-1)}(\mathbf{x}_g)$ explores the significant density region of $p^{(i)}(\mathbf{x}_g)$, the adjacent intermediate densities $p^{(i-1)}(\mathbf{x}_g)$ and $p^{(i)}(\mathbf{x}_g)$ cannot be too different. This difference is described by the estimated COV of a relative weight in terms of the samples from the Gaussian mixture density. Denote this estimated COV as $\hat{\delta}_{\omega_r}$, and see Appendix 1 for details of the derivation of it. When $p^{(i-1)}(\mathbf{x}_g)$ is determined, the choice of $p^{(i)}(\mathbf{x}_g)$ should meet the requirement that $\hat{\delta}_{\omega_r} \leq \delta_{\omega_r,\text{thr}}$, where $\delta_{\omega_r,\text{thr}}$ is the threshold value of $\hat{\delta}_{\omega_r}$. $\delta_{\omega_r,\text{thr}}$ cannot be too large or too small and is taken as 100% in this paper according to the previous similar research [22].

As can be seen from Eq. (11), $p^{(i)}(\mathbf{x}_g)$ completely depends on $\eta^{(i)}(\mathbf{x}_g)$. Therefore, only the expression of $\eta^{(i)}(\mathbf{x}_g)$ is given in the following. Consider that [33]

$$I\left[g(\mathbf{x}_g) \leq 0\right] \simeq \lim_{\kappa \to 0^+} \Phi\left[-\frac{g(\mathbf{x}_g)}{\kappa}\right], \tag{12}$$

the formulations of $\eta^{(i)}(\mathbf{x}_g)$ for estimating $I_1$ and $I_2$ are respectively taken as

$$\eta_1^{(i_1)}(\mathbf{x}_g) = \Phi\left[-\frac{g(\mathbf{x}_g)}{\kappa^{(i_1)}}\right] L(\mathbf{x}_g)^{\lambda_1^{(i_1)}} \pi(\mathbf{x}_g), i_1 = 1, 2, ..., M_1 - 1, \tag{13}$$

$$\eta_2^{(i_2)}(\mathbf{x}_g) = L(\mathbf{x}_g)^{\lambda_2^{(i_2)}} \pi(\mathbf{x}_g), i_2 = 1, 2, ..., M_2 - 1, \tag{14}$$

where $\kappa$ and $\lambda_{1,2}$ are auxiliary parameters, $M_1$ and $M_2$ are the numbers of intermediate densities, $+\infty > \kappa^{(1)} > \kappa^{(2)} > ... \geq \kappa^{(M_1-1)} \geq 0$ (here $\kappa = 0$ represents $\Phi\left[-\frac{g(\mathbf{x}_g)}{\kappa}\right] = I\left[g(\mathbf{x}_g) \leq 0\right]$), $0 < \lambda_1^{(1)} < \lambda_1^{(2)} < ... \leq \lambda_1^{(M_1-1)} \leq 1$ and $0 < \lambda_2^{(1)} < \lambda_2^{(2)} < ... < \lambda_2^{(M_2-1)} < 1$. For the convenience of illustration, take the estimation of $I_1$ as an example. Given a $\eta_1^{(i_1-1)}(x_g)$ parameterized by $\kappa^{(i_1-1)}$ and $\lambda_1^{(i_1-1)}$, one needs to search for $\kappa^{(i_1)}$ and $\lambda_1^{(i_1)}$ that satisfy $\hat{\delta}_{\omega_r} = \delta_{\omega_r,\text{thr}}$, where $\hat{\delta}_{\omega_r}$ is estimated with the samples from $p_{1G}^{(i_1-1)}(x_g)$. Note that there are infinite combinations of $\kappa^{(i_1)}$ and $\lambda_1^{(i_1)}$ that satisfy this condition, the strategy in this paper is to first determine $\kappa^{(i_1)}$ such that $\hat{\delta}_{\omega_r} = \delta_{\omega_r,\text{thr}}/2$ holds and then find $\lambda_1^{(i_1)}$. An issue exists that whether $\kappa$ reaches 0 first or $\lambda_1$ reaches 1 first. In the first case, $\kappa^{(i_1)} = 0$ and $\lambda_1^{(i_1)} < 1$, then from step $i_1+1$, $\kappa$ remains equal to 0 and only $\lambda_1$ is increased. Otherwise $\kappa^{(i_1)} > 0$ and $\lambda_1^{(i_1)} = 1$, then from step $i_1+1$, only $\kappa$ is decreased and $\lambda_1$ remains equal to 1. The latter situation is more likely to occur than the former, since $\hat{\delta}_{\omega_r}$ is more sensitive to the change of the value of $\kappa$. However, this issue does not exist for estimating $I_2$ because only one parameter $\lambda_2$ is involved.

*3.2. Construction scheme of Gaussian mixture density*

It is indicated in section 3.1 that the number of Gaussian mixture densities equals to that of intermediate densities. This section explains how to construct the Gaussian mixture density of each intermediate step based on the intermediate density of the current step and the samples from the Gaussian mixture density of the previous step. A standard Gaussian mixture density can be described as

$$p_G(x_g) = \sum_{j=1}^{K} \pi_j \psi(x_g \mid \mu_j, \Sigma_j), \tag{15}$$

where $K$ is the number of Gaussian densities, $\pi_j$, $\mu_j$ and $\Sigma_j$ are respectively the weight coefficient, mean vector and covariance matrix of the $j$-th density, $0 \leq \pi_j \leq 1$, $\sum_{j=1}^{K} \pi_j = 1$, and $\psi(\cdot)$ is the PDF of Gaussian type. The construction of a Gaussian mixture density equals to the determination of $K$, $\pi_j$, $\mu_j$ and $\Sigma_j$ ($j = 1, 2, ..., K$). Given a $p_G^{(i-1)}(x_g)$ and a $p^{(i)}(x_g)$, the construction scheme of $p_G^{(i)}(x_g)$ is summarized in Algorithm 1. Although theoretically we need to compute the weights of the samples with respect to $p^{(i)}(x_g)$ in step 1, given that only the relative magnitude of these weights matters, not the absolute one, the weights with respect to $\eta^{(i)}(x_g)$ is used here. Note that one needs to choose an appropriate $K$ in step 2. If it is too small, the Gaussian mixture density may not cover the significant density region of the intermediate density; if too large, the computational cost will be too high because $N_G$ should increase accordingly to ensure that each component Gaussian density is sampled. It is suggested that $K$ is slightly larger than the maximum of the three numbers: dimension, subsystem and observation of the problem [25], and $N_G$ is around 50 times the value of $K$. Moreover, the choice of $\Sigma_j$ in step 5 is conservative, where the minimization function is used to cope with multiple cases such as unimodal and multimodal target densities. The covariance matrix is large enough to ensure that the tails of $p_G^{(i)}(x_g)$ are not "thinner" than that of $p^{(i)}(x_g)$ such that samples from $p_G^{(i)}(x_g)$ can effectively explore the significant density regions of $p^{(i)}(x_g)$ and the next intermediate density, $p^{(i+1)}(x_g)$, as well.

**Algorithm 1.** Construction scheme of Gaussian mixture density

1. Generate $N_G$ samples from $p_G^{(i-1)}(x_g)$, denoted as $\{x_g^{(k)} : k = 1, 2, ..., N_G\}$, and compute the weights of these samples with respect to $\eta^{(i)}(x_g)$;
2. Set a value for $K$, and cluster the samples in step 1 into $K$ clusters using $K$-means clustering algorithm. Note that each sample $x_g^{(k)}$ is with a weight
$$\omega_2^{(k)} = \eta^{(i)}(x_g^{(k)}) / p_G^{(i-1)}(x_g^{(k)});$$
3. Take the mass center of the $j$-th cluster as $\mu_j$, and set the weight coefficients are equal, i.e., $\pi_j = 1/K$ ($j = 1, 2, ..., K$);
4. Estimate the variance $\hat{\sigma}^2$ using the samples in step 1, the $i$-th component of $\hat{\sigma}^2$ is derived as $\hat{\sigma}_i^2 = \sum_{k=1}^{N_G} \bar{\omega}_2^{(k)} (x_{gi}^{(k)} - \hat{\mu}_i)^2$, where $x_{gi}^{(k)}$ is the $i$-th component of $x_g^{(k)}$, $\bar{\omega}_2^{(k)} = \omega_2^{(k)} / \sum_{k=1}^{N_G} \omega_2^{(k)}$ is the $k$-th normalized weight, $\hat{\mu}_i$ is the $i$-th component of $\hat{\mu}$, the mass center of all samples;
5. Let all the $K$ Gaussian densities use a same covariance matrix, $\Sigma_j = \text{diag}\left[\min(\hat{\sigma}_1^2, 1), \min(\hat{\sigma}_2^2, 1), ..., \min(\hat{\sigma}_{n_g}^2, 1)\right] (j = 1, 2, ..., K)$, where $\text{diag}(v)$ denotes the diagonal matrix containing the elements of the vector $v$ on the main diagonal.

*3.3. Update the Gaussian mixture density based on cross entropy method*

In the previous two sections, the determination methods of the intermediate densities and Gaussian mixture densities are introduced. Starting from $\pi(x_g)$ (here $\pi(x_g)$ is both $p_G^{(0)}(x_g)$ and $p^{(0)}(x_g)$) together with its $N_G$ samples, the first intermediate density $p^{(1)}(x_g)$ is determined with parameters $\kappa^{(1)}$ and $\lambda_1^{(1)}$ (for estimating $I_1$), or parameter $\lambda_2^{(1)}$ (for estimating $I_2$) by constraining the estimated COV of the relative weight. Subsequently, the first Gaussian mixture density $p_G^{(1)}(x_g)$ is constructed using Algorithm 1. Based on $p^{(1)}(x_g)$ and $N_G$ samples from $p_G^{(1)}(x_g)$, the second intermediate density $p^{(2)}(x_g)$ is determined next. Repeat the process above and one obtains a sequence of intermediate densities and Gaussian mixture densities. For estimating $I_1$, this procedure is terminated and let $M_1$ be equal to the current $i_1$ plus 1 when both of the following conditions are satisfied: (1) $\lambda_1^{(i_1)} = 1$; (2) $\sum_{k=1}^{N_G} I\left[g\left(x_g^{(k)}\right) \leq 0\right]/N_G \geq I_{\mathrm{f,thr}}$, where $I_{\mathrm{f,thr}}$ is the threshold value for the ratio of the number of samples falling into the failure domain to the total number of samples and is suggested to be 0.1 according to experimental study. For estimating $I_2$, however, only one condition $\lambda_2^{(i_2)} = 1$ needs to be satisfied, and $M_2$ equals to the current $i_2$ when this condition holds.

Note that the Gaussian mixture density in step $M_1-1$ (or $M_2$) is only a rough approximation of the optimal ISD, and there is still a certain gap with the optimal ISD. Therefore, this Gaussian mixture density is then adaptively modified through the cross entropy method. As an efficient measure of the difference between two probability densities, the cross entropy between $p_G(x_g)$ and $p_{\mathrm{opt}}(x_g)$ is defined as

$$H\left[p_{\mathrm{opt}}(x_g), p_G(x_g)\right] = -\int_{x_g} p_{\mathrm{opt}}(x_g) \ln p_G(x_g) \mathrm{d}x_g . \tag{16}$$

Here this cross entropy is reduced using an iteration algorithm proposed by Kurtz and Song [25]. Given a Gaussian mixture density indicated in Eq. (15) and $N$ samples from it, the mean vector, covariance matrix and the weight coefficient of the $j$-th ($j = 1, 2,…, K$) component density of the next Gaussian mixture density, denoted as $\mu'_j, \Sigma'_j$ and $\pi'_j$, are respectively derived as

$$\mu'_j = \frac{\sum_{k=1}^{N} \gamma_j^{(k)} x_g^{(k)} \eta^{(M)}\left(x_g^{(k)}\right)/p_G\left(x_g^{(k)}\right)}{\sum_{k=1}^{N} \gamma_j^{(k)} \eta^{(M)}\left(x_g^{(k)}\right)/p_G\left(x_g^{(k)}\right)} , \tag{17}$$

$$\Sigma'_j = \frac{\sum_{k=1}^{N} \gamma_j^{(k)} \left[ x_g^{(k)} - \mu_j \right] \left[ x_g^{(k)} - \mu_j \right]^{\mathrm{T}} \eta^{(M)}\left(x_g^{(k)}\right) \Big/ p_G\left(x_g^{(k)}\right)}{\sum_{k=1}^{N} \gamma_j^{(k)} \eta^{(M)}\left(x_g^{(k)}\right) \Big/ p_G\left(x_g^{(k)}\right)}, \tag{18}$$

$$\pi'_j = \frac{\sum_{k=1}^{N} \gamma_j^{(k)} \eta^{(M)}\left(x_g^{(k)}\right) \Big/ p_G\left(x_g^{(k)}\right)}{\sum_{k=1}^{N} \eta^{(M)}\left(x_g^{(k)}\right) \Big/ p_G\left(x_g^{(k)}\right)}, \tag{19}$$

where

$$\gamma_j^{(k)} = \frac{\pi_j \psi\left(x_g^{(k)} \mid \mu_j, \Sigma_j\right)}{\sum_{l=1}^{K} \pi_l \psi\left(x_g^{(k)} \mid \mu_l, \Sigma_l\right)}. \tag{20}$$

The iteration algorithm above is repeated until the stopping criterion is satisfied. Here a condition in terms of the COV of the estimator is given in the following. Take $\hat{I}_1$ as an example, the standard deviation of $\hat{I}_1$ can be estimated using the $N_1$ samples from the Gaussian mixture density $p_{1G}(x_g)$ according to Eq. (22) (will be presented in section 4), i.e.,

$$\hat{\sigma}_1 = \sqrt{\frac{1}{N_1} \left[ \frac{1}{N_1} \sum_{k=1}^{N_1} \left\{ \frac{I\left[g\left(x_g^{(k)}\right) \leq 0\right] L\left(x_g^{(k)}\right) \pi\left(x_g^{(k)}\right)}{p_{1G}\left(x_g^{(k)}\right)} \right\}^2 - \hat{I}_1^2 \right]}. \tag{21}$$

The estimated COV of $\hat{I}_1$ is thus calculated by $\hat{\delta}_1 = \hat{\sigma}_1 / \hat{I}_1$. The cross entropy-based iteration procedure is terminated and the updated Gaussian mixture density is chosen as the final ISD when $\hat{\delta}_1 \leq \delta_{1,\mathrm{thr}}$, where $\delta_{1,\mathrm{thr}}$ is the threshold value of $\hat{\delta}_1$. This stopping criterion is analogous for $\hat{I}_2$. Moreover, both threshold COVs of $\hat{I}_1$ and $\hat{I}_2$ are taken 5% in this paper.

*3.4. Summarize of RU-SAIS*

To make the steps described in sections 3.1-3.3 more visually, a flowchart is presented in Fig. 1. The detailed algorithm is summarized in Algorithm 2 accordingly. It is noted that the number of samples (i.e., the number of likelihood function calls) required in the whole algorithm equals to $N_G + N_G(M_1-2) + N_1(N_{\mathrm{CE},1}+1) + N_G(M_2-1) + N_2(N_{\mathrm{CE},2}+1)$ = $N_G(M_1+M_2-2) + N_1(N_{\mathrm{CE},1}+1) + N_2(N_{\mathrm{CE},2}+1)$, where $N_{\mathrm{CE},1}$ and $N_{\mathrm{CE},2}$ are respectively the numbers of runs of the cross entropy-based iteration algorithm for estimating $I_1$ and $I_2$. This will be clarified in the following examples. Moreover, considering the limit state function $g(\boldsymbol{x}_g)$ is only appeared in the expression of $I_1$, the number of limit state function calls is thus $N_G(M_1-1) + N_1(N_{CE,1}+1)$.

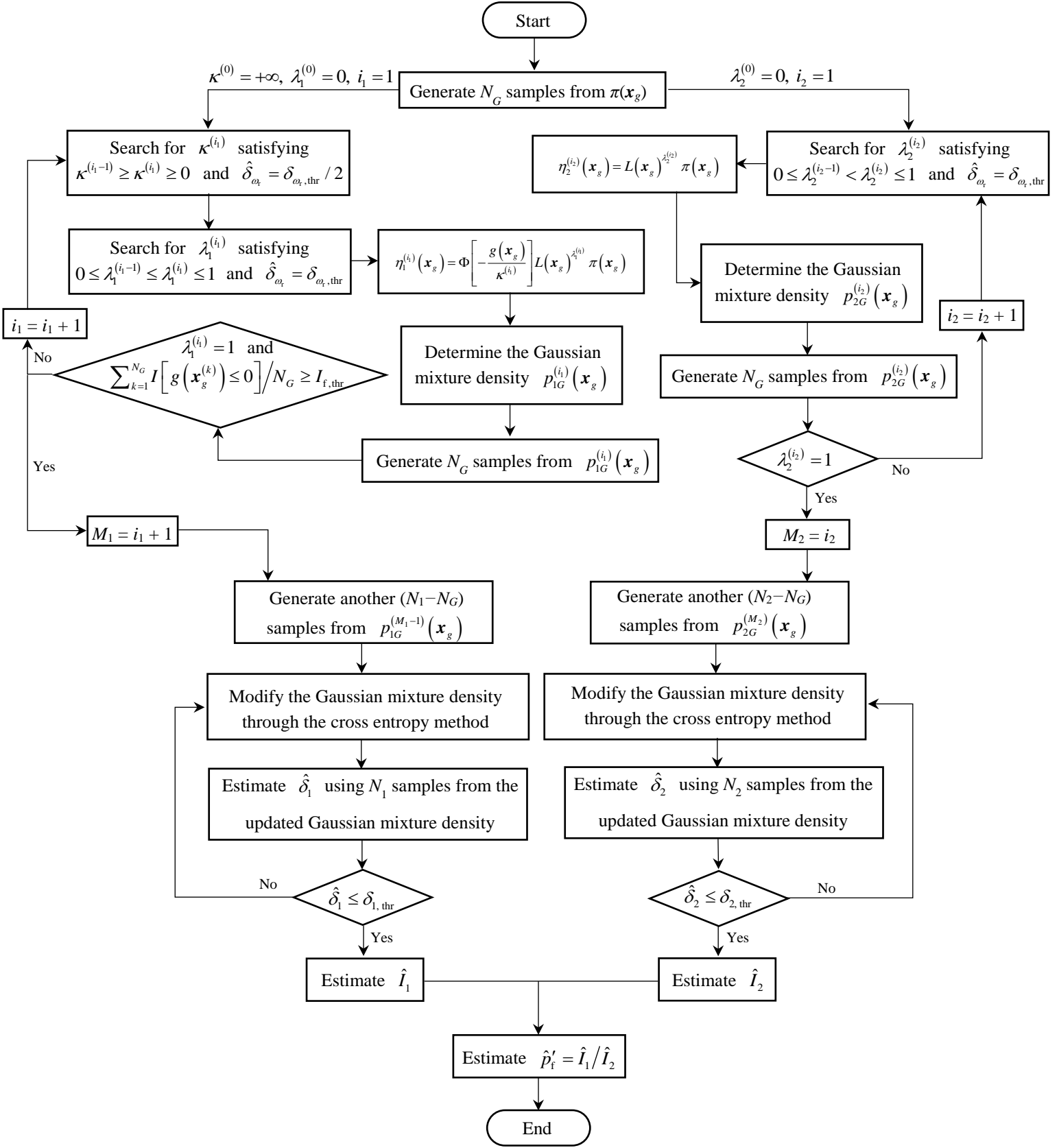

**Fig. 1.** Flowchart of RU-SAIS

**Algorithm 2.** Reliability updating using sequential adaptive importance sampling (RU-SAIS)

0. Generate $N_G$ samples from $\pi(x_g)$;

***Sub-algorithm 1.*** *Estimation of* $\hat{I}_1$

PART I: sequential adaptive importance sampling

1.1. Set $\kappa^{(0)} = +\infty$, $\lambda_1^{(0)} = 0$, $i_1 = 1$;

1.2. If $\lambda_1^{(i_1-1)} = 1$, set $\lambda_1^{(i_1)} = 1$, and search for $\kappa^{(i_1)}$ satisfying $\kappa^{(i_1-1)} \geq \kappa^{(i_1)} \geq 0$ and $\hat{\delta}_{\omega_r} = \delta_{\omega_r,\text{thr}}$, then go to step 1.4; otherwise, search for $\kappa^{(i_1)}$ satisfying $\kappa^{(i_1-1)} \geq \kappa^{(i_1)} \geq 0$ and $\hat{\delta}_{\omega_r} = \delta_{\omega_r,\text{thr}}/2$, then go to step 1.3;

1.3. Search for $\lambda_1^{(i_1)}$ satisfying $0 \leq \lambda_1^{(i_1-1)} \leq \lambda_1^{(i_1)} \leq 1$ and $\hat{\delta}_{\omega_r} = \delta_{\omega_r,\text{thr}}$;

1.4. Substitute $\kappa^{(i_1)}$ and $\lambda_1^{(i_1)}$ into the intermediate density

$$\eta_1^{(i_1)}(x_g) = \Phi\left[-\frac{g(x_g)}{\kappa^{(i_1)}}\right] L(x_g)^{\lambda_1^{(i_1)}} \pi(x_g);$$

1.5. Determine the Gaussian mixture density $p_{1G}^{(i_1)}(x_g)$ according to Algorithm 1, and generate $N_G$ samples from it;

1.6. Check whether both of the following conditions are satisfied: (i) $\lambda_1^{(i_1)} = 1$; (2) $\sum_{k=1}^{N_G} I\left[g(x_g^{(k)}) \leq 0\right]/N_G \geq I_{f,\text{thr}}$. If so, set $M_1 = i_1 + 1$, then go to step 1.7; if not, set $i_1 = i_1 + 1$, then return to step 1.2.

PART II: cross entropy-based iteration

1.7. Generate another ($N_1$-$N_G$) samples from $p_{1G}^{(M_1-1)}(x_g)$;

1.8. Modify the Gaussian mixture density through the cross entropy method as described in section 3.3;

1.9. Estimate $\hat{\delta}_1$ using $N_1$ samples from the updated Gaussian mixture density;

1.10. Check whether $\hat{\delta}_1 \leq \delta_{1,\text{thr}}$. If so, estimate $\hat{I}_1$; if not, return to step 1.8.

***Sub-algorithm 2.*** *Estimation of* $\hat{I}_2$

PART I: sequential adaptive importance sampling

2.1. Set $\lambda_2^{(0)} = 0$, $i_2 = 1$;

2.2. Search for $\lambda_2^{(i_2)}$ satisfying $0 \leq \lambda_2^{(i_2-1)} < \lambda_2^{(i_2)} \leq 1$ and $\hat{\delta}_{\omega_r} = \delta_{\omega_r,\text{thr}}$;

2.3. Substitute $\lambda_2^{(i_2)}$ into the intermediate density $\eta_2^{(i_2)}(x_g) = L(x_g)^{\lambda_2^{(i_2)}} \pi(x_g)$;

2.4. Determine the Gaussian mixture density $p_{2G}^{(i_2)}(x_g)$ according to Algorithm 1, and generate $N_G$ samples from it;

2.5. Check whether $\lambda_2^{(i_2)} = 1$. If so, set $M_2 = i_2$, then go to step 2.6; if not, set $i_2 = i_2 + 1$, then return to step 2.2.

PART II: cross entropy-based iteration

2.6. Generate another ($N_2$-$N_G$) samples from $p_{2G}^{(M_2)}(x_g)$;

2.7. Modify the Gaussian mixture density through the cross entropy method as described in section 3.3;

2.8. Estimate $\hat{\delta}_2$ using $N_2$ samples from the updated Gaussian mixture density;

2.9. Check whether $\hat{\delta}_2 \leq \delta_{2,\text{thr}}$. If so, estimate $\hat{I}_2$; if not, return to step 2.7.

3. Estimate $p'_f = \hat{I}_1 / \hat{I}_2$.

## 4. Statistical properties of the estimator

In this section, the statistical properties including the expectation and COV of the estimator indicated in Eq. (8) are derived. Since $I_1$ and $I_2$ are both estimated using IS, only the derivation of the statistical properties of $\hat{I}_1$ is given here, and the statistical properties of $\hat{I}_2$ can be obtained in the same way. Under the assumption that the support of $p_1(x_g)$ contains that of $p_{1,\text{opt}}(x_g)$, $\hat{I}_1$ is an unbiased estimator of $I_1$, i.e.,

$E_{p_1}(\hat{I}_1) = I_1$, and the variance of $\hat{I}_1$ is given by [24]

$$Var_{p_1}(\hat{I}_1) = \frac{1}{N_1}\left[\int_{x_g}\left\{\frac{I[g(x_g)\leq 0]L(x_g)\pi(x_g)}{p_1(x_g)}\right\}^2 p_1(x_g)dx_g - I_1^2\right], \quad (22)$$

where $E_{p_1}$ and $Var_{p_1}$ denote respectively the expectation and variance with respect to $p_1(x_g)$. According to the central limit theorem, $\hat{I}_1$ will be approximately normally distributed with mean $I_1$ and standard deviation $\sigma_1 = \sqrt{Var_{p_1}(\hat{I}_1)}$ if $N_1$ is sufficiently large. In the same way, when $N_2$ is sufficiently large, $\hat{I}_2$ will be approximately normally distributed with mean $I_2$ and standard deviation $\sigma_2 = \sqrt{Var_{p_2}(\hat{I}_2)}$ where

$$Var_{p_2}(\hat{I}_2) = \frac{1}{N_2}\left[\int_{x_g}\left\{\frac{L(x_g)\pi(x_g)}{p_2(x_g)}\right\}^2 p_2(x_g)dx_g - I_2^2\right]. \quad (23)$$

In this context, $\hat{p}'_f$ is approximately the quotient of two independent normally distributed variables, i.e.,

$$\hat{p}'_f \simeq \frac{\sigma_1 U_1 + I_1}{\sigma_2 U_2 + I_2} = \frac{I_1}{I_2} \cdot \frac{\sigma_1 U_1/I_1 + 1}{\sigma_2 U_2/I_2 + 1}. \quad (24)$$

Toward the goal of obtaining the statistical properties of $\hat{p}'_f$, we consider a variable $T_1 = (aU_1 + 1)/(bU_2 + 1)$, where $a$, $b$ are positive constants. See Appendix 2 for the derivation of the PDF of $T_1$. In Eq. (24), $a$ and $b$ respectively equal to the COVs of $\hat{I}_1$ and $\hat{I}_2$ which are both controlled within a threshold value. Considering the situation where $a = b = 5\%$, the PDF curve of $T_1$ can be plotted as shown in Fig. 2.

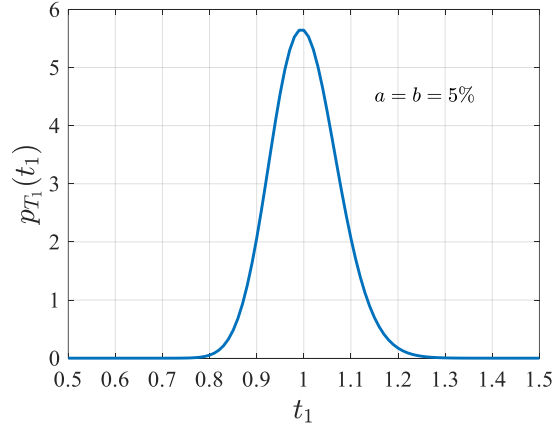

**Fig. 2.** The PDF of $T_1$ when $a = b = 5\%$

It is indicated in Fig. 2 that $T_1$ is asymmetrically distributed and slightly skew to the right. However, the skewness of this distribution is very small and will be further reduced if the COVs of $\hat{I}_1$ and $\hat{I}_2$ are controlled within a smaller value. Considering that the mean and standard deviation of $T_1$ are intractable to solve due to the sophisticated PDF, here they are both estimated through an additional simple sampling, i.e.,

$$\hat{\mu}_{T_1} = \frac{1}{N_{T_1}} \sum_{k=1}^{N_{T_1}} \frac{au_1^{(k)}+1}{bu_2^{(k)}+1}, \tag{25}$$

$$\hat{\sigma}_{T_1} = \sqrt{\frac{1}{N_{T_1}-1} \sum_{k=1}^{N_{T_1}} \left( \frac{au_1^{(k)}+1}{bu_2^{(k)}+1} - \hat{\mu}_{T_1} \right)^2}, \tag{26}$$

where $u_1^{(k)}$ and $u_2^{(k)}$ are respectively the $k$-th samples of $U_1$ and $U_2$, and $N_{T_1}$ is the number of joint samples. When $a = b = 5\%$, set $N_{T_1} = 10^6$, and one can obtain that the estimated mean and standard deviation are respectively 1.0025 and 0.0712. In this context, $\dfrac{\sigma_1 U_1 + I_1}{\sigma_2 U_2 + I_2}$ is a biased estimator for the posterior failure probability but the bias is very small. Moreover, the COV of $\hat{p}'_f$ can be approximately estimated by $\hat{\delta}_{\hat{p}'_f} = \hat{\sigma}_{T_1}/\hat{\mu}_{T_1} = 0.071$, which indicates that the COV of $\hat{p}'_f$ can be controlled within 7% if both COVs of $\hat{I}_1$ and $\hat{I}_2$ are controlled within 5%.

## 5. Example

In this section, three examples are presented in depth to demonstrate the computational efficiency of the proposed method. Taking the results of crude Monte Carlo simulation (MCS) as the benchmark, both the results of RU-SAIS and subset simulation-based reliability updating method are presented to make a comparison. The first two-dimensional numerical example is designed to illustrate the detailed process of RU-SAIS and the computational capacity for reliability updating problems where the posterior PDF is multimodal. In the second example, a truss system involving ten random variables is elaborated to demonstrate the applicability to high-dimensional problems. This is a typical dynamic updating problem as the information is obtained in sequence. Finally, the proposed method is implemented to an engineer-guided concrete corrosion case. It is noted that different types of observations are used in these examples. These include the measurements of the random variables themselves (in example 2) and those of the response functions (in examples 1 and 3).

*5.1. two-dimensional numerical case*

The first example is a two-dimensional problem involving a multimodal posterior density. The limit state function comes from [25] and writes

$$g(x_g) = 5 - x_2 - 0.5(x_1 - 0.1)^2, \tag{27}$$

where $X_g = [X_1; X_2]$ is a vector consisting of two mutually independent standard normal random variables a priori, i.e., $\pi(x_g) = \varphi(x_1)\varphi(x_2)$. Now we have two observations, one is a measurement of one response function $s_1(x_g) = x_2 - 1.4x_1$ with value of 1.6, i.e., $s_{1m} = 1.6$, and the other is a measurement of $s_2(x_g) = x_1^2 + 4.4x_1 - x_2$ with value of 2.4, i.e., $s_{2m} = 2.4$. The prediction errors, defined by $\varepsilon_1 = s_1(x_g) - s_{1m}$ and $\varepsilon_2 = s_2(x_g) - s_{2m}$, are normally distributed with means both 0 and standard deviations 0.8 and 1.0, respectively. The likelihood function can be thus taken as

$$L(x_g) = \exp\left\{-\frac{1}{2}\left[\frac{(x_2 - 1.4x_1 - 1.6)^2}{0.8^2} + \frac{(x_1^2 + 4.4x_1 - x_2 - 2.4)^2}{1.0^2}\right]\right\}. \tag{28}$$

Fig. 3 presents the limit state surface (denoted by the solid parabolic line) and two contours of functions proportional to the two optimal ISDs indicated in Eq. (9) and (10), respectively. Observe that the second density $p_{2,\text{opt}}(x_g)$, i.e., the posterior PDF, is near unimodal because the PDF value of one peak is much smaller than the other. However, the first density $p_{1,\text{opt}}(x_g)$ becomes a bimodal one due to the shape of the limit state surface. Toward the goal of obtaining a near-optimal ISD resembling $p_{1,\text{opt}}(x_g)$, according to RU-SAIS, we start with the prior PDF $\pi(x_g)$ and draw $N_G = 500$ samples from it. Using these samples, we first search for a $\kappa^{(1)}$ to make the estimated COV of the relative weight, $\hat{\delta}_{\omega_r}$, equals to $\delta_{\omega_r,\text{thr}}/2 = 50\%$. In a particular run of the proposed method, $\kappa^{(1)} = 4.24$. Fix this value of $\kappa^{(1)}$, and then we search for the $\lambda_1^{(1)}$ satisfying $\hat{\delta}_{\omega_r} = 100\%$ using the same samples. We obtain $\lambda_1^{(1)} = 0.13$ and thus we get the first intermediate density determined by

$$\eta_1^{(1)}(x_g) = \Phi\left[-\frac{g(x_g)}{4.24}\right] L(x_g)^{0.13} \pi(x_g). \tag{29}$$

Now we can compute the weights of the 500 samples with respect to $\eta_1^{(1)}(x_g)$ and these weighted samples are clustered with $K$-means clustering algorithm. Set $K = 10$, and the clustering result and the mass center of each cluster are presented in Fig. 4. Moreover, the estimated variance of these samples is $\hat{\sigma}^2 = [0.37; 0.94]$. According to Algorithm 1, the first Gaussian mixture density $p_{1G}^{(1)}(x_g)$ is constructed using 10 component Gaussian densities, each with a weight coefficients of 1/10, a covariance matrix with diag(0.37,0.94) and a mean vector corresponding to the mass center of each cluster. Draw $N_G = 500$ samples from $p_{1G}^{(1)}(x_g)$ and repeat the foregoing procedures, and the contours of functions proportional to the intermediate density and the corresponding Gaussian mixture density in each step are elaborated in Fig. 5.

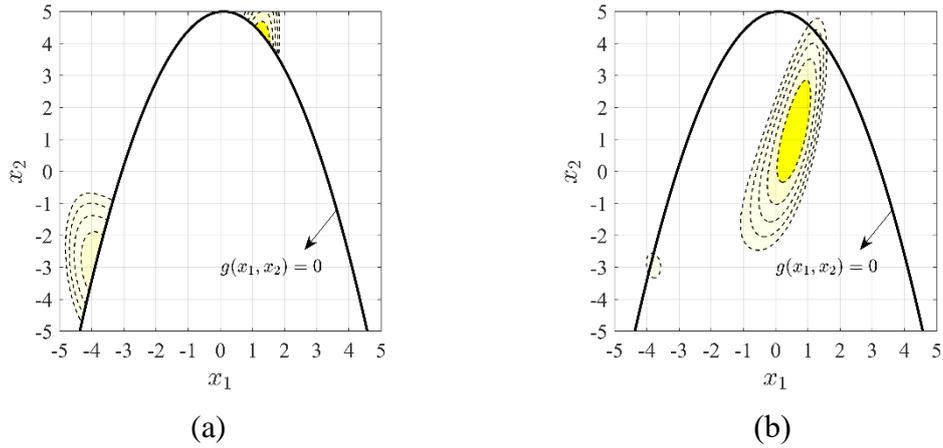

**Fig. 3.** Limit state surface and contours of functions proportional to the optimal ISDs for estimating (a) $I_1$ and (b) $I_2$; the depth of color indicates the size of the probability density, and the higher the probability density, the darker the color

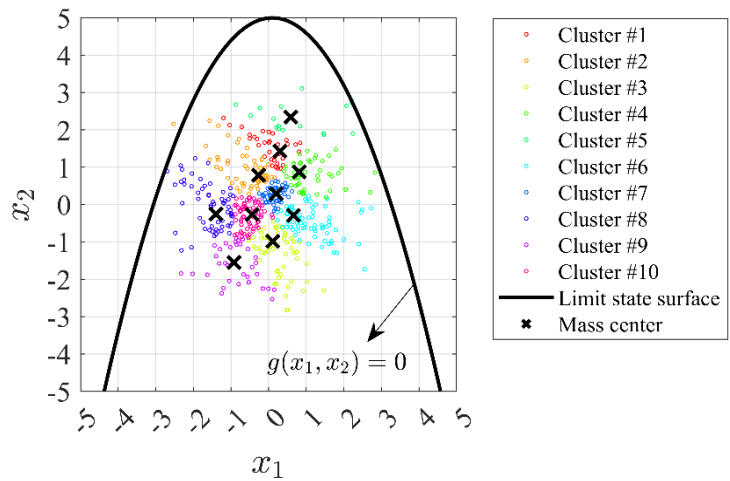

**Fig. 4.** Clustering results of the weighted samples using *K*-means clustering algorithm and the mass center of each cluster

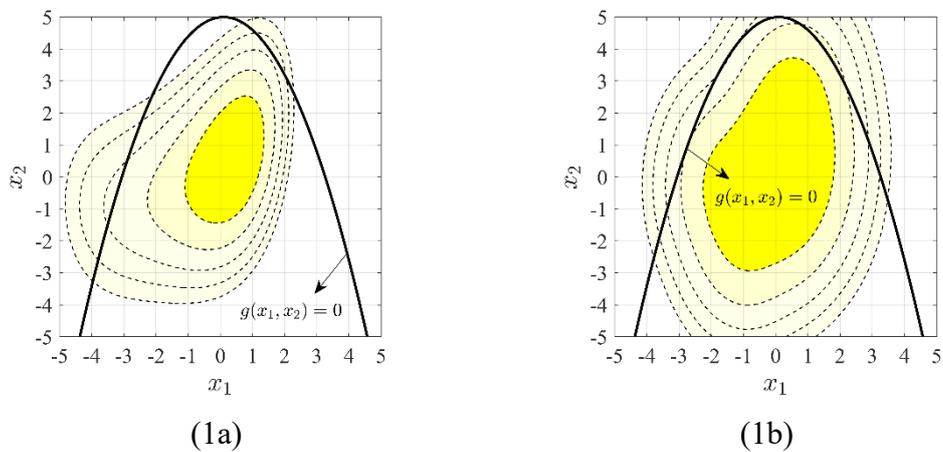

(1a)            (1b)

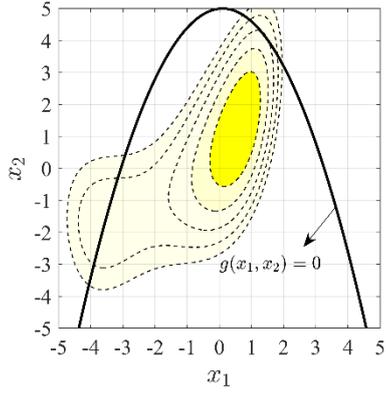
(2a)

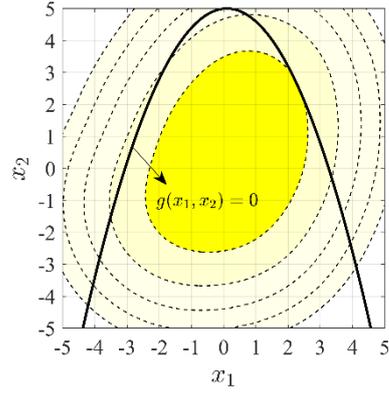
(2b)

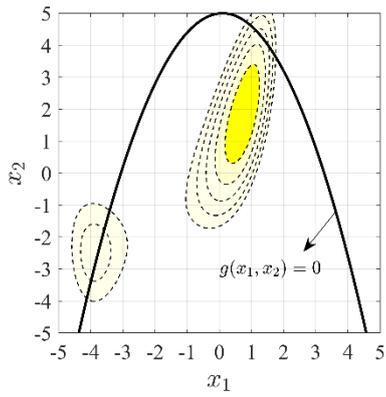
(3a)

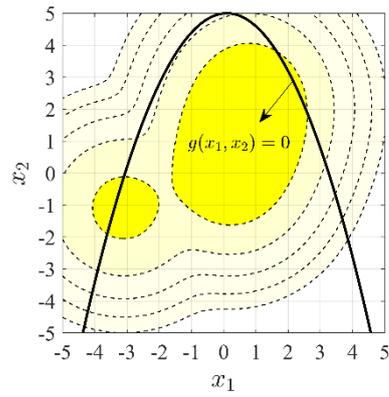
(3b)

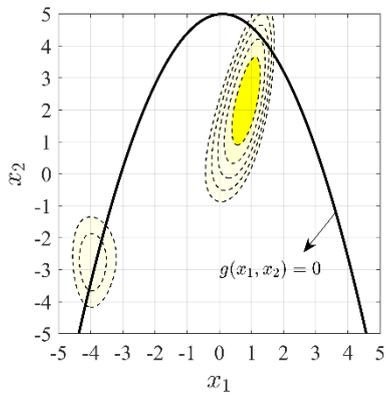
(4a)

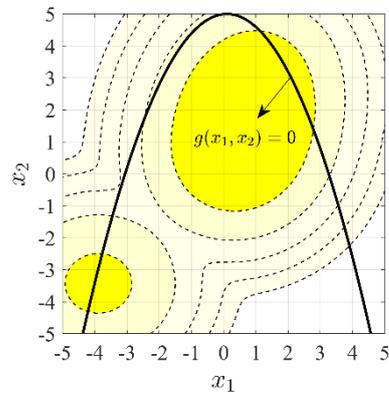
(4b)

**Fig.5.** Contours of functions proportional to: (1a) the intermediate density of step 1 ( $\kappa^{(1)} = 4.24$, $\lambda_1^{(1)} = 0.13$ ); (1b) the Gaussian mixture density of step 1; (2a) the intermediate density of step 2 ( $\kappa^{(2)} = 2.40$, $\lambda_1^{(2)} = 0.33$ ); (2b) the Gaussian mixture density of step 2 ( $\sum_{k=1}^{N_G} I\left[ g\left(x_g^{(k)}\right) \leq 0 \right] \Big/ N_G = 0.03$ ); (3a) the intermediate density of step 3 ( $\kappa^{(3)} = 1.76$, $\lambda_1^{(3)} = 0.71$ ); (3b) the Gaussian mixture density of step 3

( $\sum_{k=1}^{N_G} I\left[ g\left(x_g^{(k)}\right) \leq 0 \right] / N_G = 0.07$ ); (4a) the intermediate density of step 4 ( $\kappa^{(4)} = 1.36$, $\lambda_1^{(4)} = 1$ ); (4b) the Gaussian mixture density of step 4 ($\sum_{k=1}^{N_G} I\left[ g\left(x_g^{(k)}\right) \leq 0 \right] / N_G = 0.11$).

One can observe from Fig. 5 that the Gaussian mixture density gradually turns from unimodal to bimodal through the sequential adaptive importance sampling procedure. Moreover, the Gaussian mixture density in each intermediate step successfully achieves a larger spread than the corresponding intermediate density so that the samples from it can explore the significant density region of the next intermediate density. In step 4, the stopping criteria $\lambda_1^{(4)} = 1$ and $\sum_{k=1}^{N_G} I\left[ g\left(x_g^{(k)}\right) \leq 0 \right] / N_G > 0.1$ are both satisfied. Therefore, $M_1$ is set as 5 and $p_{1G}^{(4)}(x_g)$ is used to conduct further cross entropy-based iteration algorithm.

Set $N_1 = 1000$, and draw another $N_1 - N_G = 500$ samples from $p_{1G}^{(4)}(x_g)$. Based on these samples, the estimated COV of $\hat{I}_1$, $\hat{\delta}_1$, equals to 56.0%, much larger than the target 5%. Fig. 6 shows the process of modifying $p_{1G}^{(4)}(x_g)$ through the cross entropy method. It is indicated that $p_{1G}^{(4)}(x_g)$ converges very fast and the estimated COV sharply decreases below 5% after only two runs of the cross entropy–based updating algorithm. The final ISD $p_1(x_g)$ indicated in Fig. 6b is very close to the optimal one indicated in Fig. 3a. $I_1$ is estimated as $2.57 \cdot 10^{-7}$ according to this ISD.

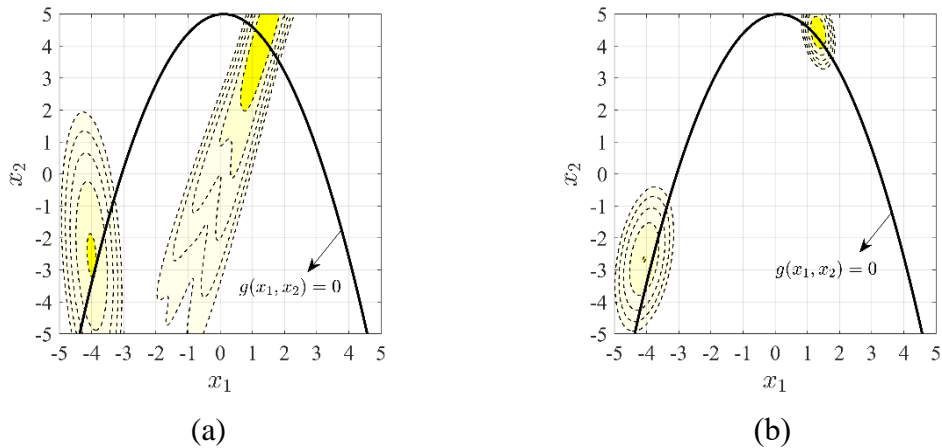

(a)                          (b)

**Fig. 6.** Contours of functions proportional to: (a) the Gaussian mixture density after one run of the cross entropy-based updating algorithm ( $\hat{\delta}_1 = 8.1\%$ ); (b) the Gaussian

mixture density after two runs of the cross entropy-based updating algorithm ($\hat{\delta}_1 = 2.7\%$).

The estimation process of $I_2$ is exactly the same as above and will not be described here. And because the optimal ISD $p_{2,\text{opt}}(x_g)$ is simpler, fewer intermediate steps are needed. Based on the same 500 samples from $\pi(x_g)$, $\lambda_2$ achieves 1 when $i_2$ equals to 3, $M_2$ is thus taken as 3. After only one run of the cross entropy-based iteration algorithm, the Gaussian mixture density $p_{2G}^{(3)}(x_g)$ satisfies the stopping criterion. Fig. 7 showcases the contours of functions proportional to $p_{2G}^{(3)}(x_g)$ and $p_2(x_g)$. $I_2$ is estimated as $1.33 \cdot 10^{-2}$ according to this ISD.

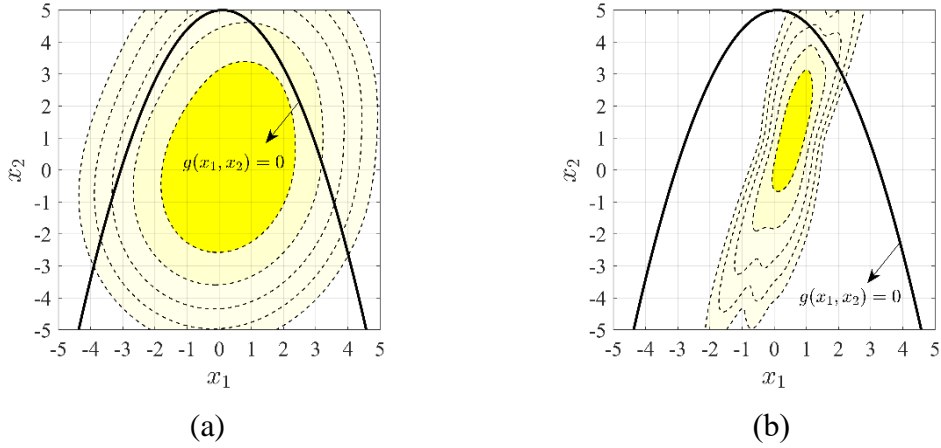

(a)          (b)

**Fig. 7.** Contours of functions proportional to: (a) $p_{2G}^{(3)}(x_g)$ ($\hat{\delta}_1 = 8.6\%$); (b) $p_2(x_g)$ ($\hat{\delta}_1 = 1.1\%$).

Table 1 compares the results of subset simulation and RU-SAIS, together with crude MCS. Subset simulation and crude MCS are used combining with Straub's method [7]. Considering that the COV of estimator obtained from crude MCS is $\sqrt{\dfrac{1-I_{1,2}}{N_{\text{MCS}}I_{1,2}}}$, the numbers of samples, $N_{\text{MCS}}$, required to achieve a 5% COV in estimating $I_1$ and $I_2$ are respectively $1.56 \cdot 10^9$ and $2.97 \cdot 10^4$. For subset simulation, suppose that the intermediate conditional probability $p_0$ equals to 0.1, and the numbers of simulation levels in estimating $I_1$ and $I_2$ are respectively 7 and 2. Given a COV of 5%, the COVs in each simulation level are approximately $\sqrt{0.05^2/7} = 1.89\%$ and $\sqrt{0.05^2/2} = 3.54\%$, respectively. According to [4], the number of samples required in each simulation level is approximately $(1+\gamma)(1-p_0)/p_0\,\delta_{i,\text{SuS}}^2$, where $1+\gamma$ is a reduction factor of the number of samples due to the sample correlation and $\delta_{i,\text{SuS}}$ is the COV in the $i$-th simulation level. Set $\gamma = 0.2$ and the numbers of samples in each simulation level are thus $3.02 \cdot 10^4$ and $8.64 \cdot 10^3$, respectively. The results in Table 1 reveal that all the COVs are close to the target 5%. At a cost of smaller number of samples, RU-SAIS achieves closer results to crude MCS than subset simulation, which implies that RU-SAIS outperforms subset simulation in terms of both accuracy and efficiency.

**Table 1.** Comparison of the results of subset simulation, RU-SAIS and crude MCS for example 1

| Method | Result of a single run | Number of samples | COV ($\hat{\delta}_1, \hat{\delta}_2$) |
|---|---|---|---|
| Crude MCS | $\dfrac{2.64 \cdot 10^{-7}}{1.36 \cdot 10^{-2}} = 1.94 \cdot 10^{-5}$ | $1.56 \cdot 10^9 + 2.97 \cdot 10^4 = 1.56 \cdot 10^9$ | 5.0%, 4.9% |
| Subset simulation | $\dfrac{4.68 \cdot 10^{-7}}{1.29 \cdot 10^{-2}} = 3.63 \cdot 10^{-5}$ | $1.92 \cdot 10^5 + 1.64 \cdot 10^4 = 2.08 \cdot 10^5$ | 9.3%, 6.5% |
| RU-SAIS | $\dfrac{2.57 \cdot 10^{-7}}{1.33 \cdot 10^{-2}} = 1.93 \cdot 10^{-5}$ | $500 \cdot 6 + 1000 \cdot 3 + 1000 \cdot 2 = 8000$ | 2.7%, 1.1% |

*5.2. Ten-dimensional truss system*

This examples follows Wang and Shafieezadeh [15]. A 23-element truss with 10 random variables is investigated here to demonstrate the capability of the proposed RU-SAIS algorithm in tackling multi-dimensional problems. Different with the original reference, the measurements in this example are obtained in sequence. As shown in Fig. 8, the structure is a simply supported truss with 11 horizontal elements and 12 diagonal ones. Six vertical concentrated forces, denoted as $\boldsymbol{P} = [P_1; P_2; \ldots; P_6]$, act on the six nodes at the top of the truss. Consider a vector consisting of 10 random variables, $[\boldsymbol{P}; E_1; A_1; E_2; A_2]$, where $E_1$ and $A_1$ are respectively the modulus of elasticity and cross section area of the diagonal elements, and $E_2$ and $A_2$ are respectively the modulus of elasticity and cross section area of the horizontal elements. Let $\boldsymbol{X}_g = [X_1; X_2; \ldots; X_{10}]$ denote the independent standard normal vector after performing an isoprobabilistic transformation to these variables. The limit state function in this example is

$$g(\boldsymbol{x}_g) = 0.14 - |d(\boldsymbol{x}_g)|, \tag{30}$$

where $d(\boldsymbol{x}_g)$ denotes the vertical displacement of the middle node at the bottom of the truss and 0.14 is the critical value of it. Assume that the involved variables are mutually independent and the prior distributions of them are given in Table 2.

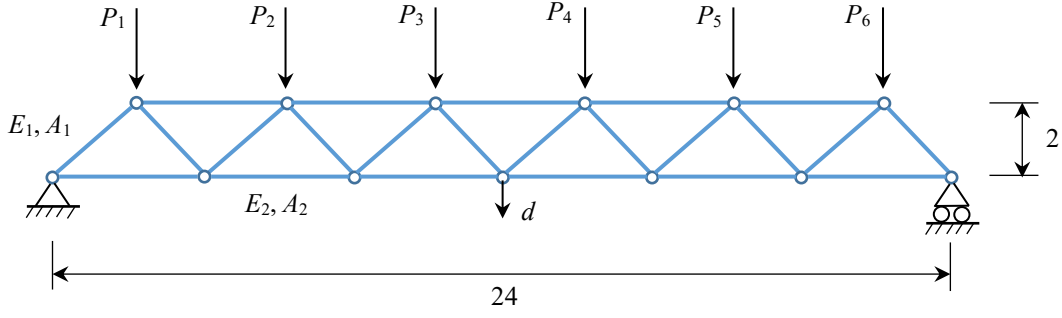

**Fig. 8.** A 23-element truss

**Table 2.** Prior distribution of random variables

| Random variable | Distribution type | Mean | Standard deviation |
|---|---|---|---|
| $P_1 \sim P_6$ | Gumbel | $6.5 \cdot 10^4$ | $6.5 \cdot 10^3$ |
| $E_1, E_2$ | Lognormal | $2.1 \cdot 10^{11}$ | $2.1 \cdot 10^{10}$ |
| $A_1$ | Lognormal | $2 \cdot 10^{-3}$ | $2 \cdot 10^{-4}$ |
| $A_2$ | Lognormal | $1 \cdot 10^{-3}$ | $1 \cdot 10^{-4}$ |

The four measurements in the original reference are assumed to be obtained in two steps in this example. The measurements of $A_1$ and $A_2$ are made first with values $1.85 \cdot 10^{-3}$ and $0.9 \cdot 10^{-3}$, respectively, and errors $\varepsilon_{1,2}$ both following a normal distribution with mean 0 and standard deviation $1 \cdot 10^{-4}$. Therefore, the likelihood function is

$$L_1(x_g) = \exp\left\{-\frac{1}{2}\left[\frac{\left(A_1(x_g) - 1.85 \cdot 10^{-3}\right)^2 + \left(A_2(x_g) - 0.9 \cdot 10^{-3}\right)^2}{\left(1 \cdot 10^{-4}\right)^2}\right]\right\}. \quad (31)$$

To obtain the posterior failure probability in this problem using RU-SAIS, we assign values for the involved parameters as following, $N_G = 1000$, $N_1 = N_2 = 2000$, $K = 20$. The results of RU-SAIS, subset simulation and crude MCS are summarized in Table 3. Here $M_1 = N_{CE,1} = 3$ and $M_2 = 2$, $N_{CE,2} = 0$. As we can see from Table 3, the results and computational costs do not vary much in these three methods because both $I_1$ and $I_2$ are not very small. Now suppose the second measurements are made, and new information is available, i.e., $P_1$ and $P_6$ have values of $8.5 \cdot 10^4$ and $7.5 \cdot 10^4$, respectively, and errors $\varepsilon_{3,4}$ both following a normal distribution with mean 0 and standard deviation $0.5 \cdot 10^4$. In this situation, the updated likelihood function is $L(x_g) = L_1(x_g) L_2(x_g)$, where

$$L_2(x_g) = \exp\left\{-\frac{1}{2}\left[\frac{\left(P_1(x_g) - 8.5 \cdot 10^4\right)^2 + \left(P_2(x_g) - 7.5 \cdot 10^4\right)^2}{\left(0.5 \cdot 10^4\right)^2}\right]\right\}. \quad (32)$$

With the new likelihood function, both crude MCS and subset simulation cannot make use of the previous results and the posterior failure probability need to be re-evaluated from scratch. However, the proposed method enables to continue the SAIS procedure based on the last Gaussian mixture density of the previous sequence. For example, for estimating the new $I_1$ with respect to $L(x_g)$, a new sequence of intermediate densities is constructed with the starting density

$$\frac{I[g(x_g) \leq 0] L_1(x_g) \pi(x_g)}{\int_{x_g} I[g(x_g) \leq 0] L_1(x_g) \pi(x_g) dx_g}, \tag{33}$$

and the final one, $p_{1,opt}(x_g)$, expressed as Eq. (9). The implementation details of the new SAIS procedure remain exactly the same, but the starting Gaussian mixture density is selected as $p_{1G}^{(2)}(x_g)$ instead of the prior density $\pi(x_g)$. According to a single run, the COV of the new estimator $\hat{I}_1$ is reduced below 5% after 4 intermediate densities and 2 runs of the cross entropy-based iteration algorithm, which requires $1000 \cdot 4 + 2000 \cdot 3 = 1 \cdot 10^4$ samples. Table 4 presents the comparison of the three methods with the additional information.

**Table 3.** Comparison of the results of subset simulation, RU-SAIS and crude MCS for example 2 when only initial information is available

| Method | Result of a single run | Number of samples | COV $(\hat{\delta}_1, \hat{\delta}_2)$ |
|---|---|---|---|
| Crude MCS | $\dfrac{1.72 \cdot 10^{-3}}{2.15 \cdot 10^{-1}} = 8.00 \cdot 10^{-3}$ | $2.20 \cdot 10^5 + 1500 = 2.22 \cdot 10^5$ | 5.1%, 4.9% |
| Subset simulation | $\dfrac{1.77 \cdot 10^{-3}}{2.21 \cdot 10^{-1}} = 8.01 \cdot 10^{-3}$ | $3.64 \cdot 10^4 + 1500 = 3.79 \cdot 10^4$ | 6.7%, 4.8% |
| RU-SAIS | $\dfrac{1.74 \cdot 10^{-3}}{2.17 \cdot 10^{-1}} = 8.02 \cdot 10^{-3}$ | $1000 \cdot 3 + 2000 \cdot 4 + 2000 \cdot 1 = 1.30 \cdot 10^4$ | 3.8%, 3.2% |

**Table 4.** Comparison of the results of subset simulation, RU-SAIS and crude MCS for example 2 when additional information is available

| Method | Result of a single run | Number of samples | COV $(\hat{\delta}_1, \hat{\delta}_2)$ |
|---|---|---|---|
| Crude MCS | $\dfrac{3.05 \cdot 10^{-5}}{2.14 \cdot 10^{-3}} = 1.43 \cdot 10^{-2}$ | $1.30 \cdot 10^7 + 2.10 \cdot 10^5 = 1.32 \cdot 10^7$ | 5.1%, 4.7% |

| Subset simulation | $\dfrac{2.49 \cdot 10^{-5}}{2.32 \cdot 10^{-3}} = 1.07 \cdot 10^{-2}$ | $9.94 \cdot 10^4 + 3.64 \cdot 10^4 = 1.36 \cdot 10^5$ | 8.2%, 6.6% |
|---|---|---|---|
| RU-SAIS | $\dfrac{3.00 \cdot 10^{-5}}{2.17 \cdot 10^{-3}} = 1.38 \cdot 10^{-2}$ | $1 \cdot 10^4 + (1000 \cdot 3 + 2000 \cdot 2) = 1.70 \cdot 10^4$ | 4.0%, 3.2% |

According to Table 4, the failure probability increases when the additional information is available, although both $I_1$ and $I_2$ get smaller. The result of RU-SAIS is still very close to that of Crude MCS, which implies the good robustness of the proposed method. By contrast, the robustness of subset simulation is reduced because the number of subsets increases. Moreover, compared with Table 3, the computational cost of RU-SAIS in Table 4 is almost the same, while it has one order of magnitude increase for subset simulation. This indicates that the computational cost of RU-SAIS is insensitive to the size of $I_1$ and $I_2$.

*5.3 Chloride and carbonation-induced concrete corrosion*

In this section, an engineer-guided case of chloride and carbonation-induced concrete corrosion is investigated to demonstrate the application of RU-SAIS in highly nonlinear problems. This example follows Xiao et al. [26]. In the original reference, the distributions of the involved five model parameters are updated based on the measurements, and the updated results are utilized to evaluate the free chloride concentration at the location of reinforcement. However, the reliability in terms of the durability limit state cannot be efficiently updated, which is the problem that this example addresses.

The following is a brief background to this example. The ingress of chloride into concrete under the influence of carbonation can be simplified as a moving boundary problem as described in Fig. 9. The carbonated zones and un-carbonated zones are separated by the reaction front. $D_{1,2}$ and $L_C$ denote the apparent chloride diffusivities of the two zones and the carbonation depth, respectively. Under the assumption that $D_{1,2}$ and $s_l$ (the pore saturation degree) are constant, Li et al. [34] derive the solution to this problem,

$$c = \frac{c_{Cl} - c_0}{c_S - c_0} = \begin{cases} A_1 \mathrm{erfc}\left(\dfrac{x}{2\sqrt{D_1 t}}\right) + A_2 & x \leq L_C = k\sqrt{t}, \\ \\ A_3 \mathrm{erfc}\left(\dfrac{x}{2\sqrt{D_2 t}}\right) & x > L_C = k\sqrt{t}. \end{cases} \quad (34)$$

with the constants

$$A_{1,2,3} = A_{1,2,3}(K_1, K_2) \text{ and } K_1 = \sqrt{\frac{D_1}{D_2}}, K_2 = \frac{k}{2\sqrt{D_2}}. \tag{35}$$

where erfc(·) denotes the complementary error function and $c$ denotes the normalized chloride concentration expressed as a function of $c_{Cl}$ (the actual chloride concentration), $c_0$ (the initial chloride concentration) and $c_S$ (the surface chloride concentration). Moreover, $k = \sqrt{2D_{CO_2}^0 \lambda_H \phi^{\lambda_1}(1-s_1)^{\lambda_2}}$ denotes the carbonation rate, where $D_{CO_2}^0 = 2 \cdot 10^{-5} \text{m}^2/\text{s}$ is the diffusivity of $CO_2$ in air, $\lambda_H$ is the carbonation rate coefficient, $\phi$ is the concrete porosity, and $\lambda_1=2.74$ and $\lambda_2=4.20$ are the exponents. The expressions of $A_{1,2,3}(K_1, K_2)$ can be referred to in [34] and are not presented here.

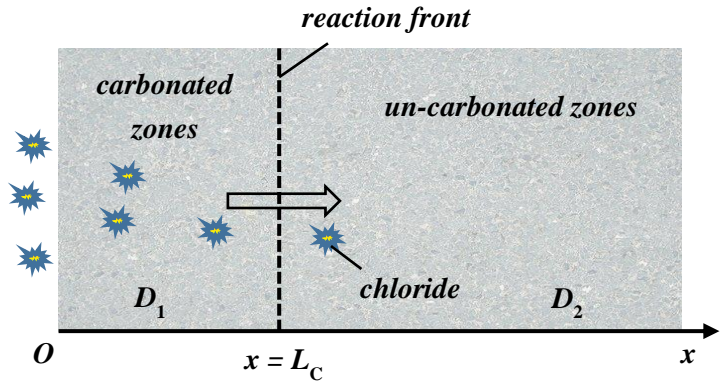

**Fig. 9.** The moving boundary problem describing the ingress of chloride into concrete under the influence of carbonation

**Table 5.** Prior distribution of random variables

| Random variable | Distribution type | Mode/Lower | Standard deviation/Upper |
| --- | --- | --- | --- |
| $\lambda_H/\lambda_{Hn}$ | Lognormal | 1.2 | 1 |
| $\phi$ | Uniform | 0.05 | 0.15 |
| $s_1$ | Uniform | 0.5 | 0.8 |
| $D_1/D_{1n}$ | Lognormal | 1 | 1 |
| $D_2/D_{2n}$ | Lognormal | 0.9 | 1 |

The involved random variables are mutually independent a priori, and the definitions and prior distributions of them are presented in Table 5, where $\lambda_{Hn} = 2 \cdot 10^{-5}$, $D_{1n} = 1 \cdot 10^{-11} \text{ m}^2/\text{s}$ and $D_{2n} = 2 \cdot 10^{-12} \text{ m}^2/\text{s}$ are the nominal values. Let $X_g = [X_1; X_2; \ldots; X_5]$ denote the independent standard normal vector after performing an isoprobabilistic transformation to these variables. Three observations at $t = 3 \cdot 10^8 \text{s}$ are taken from the original reference [26] that the normalized concentrations at 0.01m and 0.03m are 0.75 and 0.35, respectively, and the carbonation depth is 0.015m. The errors of the measured carbonation concentration and carbonation depth are both defined as the ratio of the model predictions and the measurements, and are both normally distributed with means both 1 and standard deviations 1/15 and 1/10. Therefore, the likelihood function in this example is

$$L(x_g) = \exp\left\{-\left[\frac{\left(c_1(x_g)/0.75-1\right)^2 + \left(c_2(x_g)/0.35-1\right)^2}{2\cdot(1/15)^2} + \frac{\left(L_C(x_g)/0.015-1\right)^2}{2\cdot(1/10)^2}\right]\right\}, \quad (36)$$

where $c_1(x_g)$ and $c_2(x_g)$ are the prediction functions of the chloride concentration at 0.01m and 0.03m, respectively. Suppose that the concrete cover thickness is 0.05m, the durability limit state function is

$$g(x_g) = 0.3 - c(x_g)\big|_{(x=0.05\text{m}, t=3\cdot10^8\text{s})}. \quad (37)$$

where 0.3 is the critical value of the normalized chloride concentration at the reinforcement. The example is highly nonlinear because both $L(x_g)$ and $g(x_g)$ are integrated with erfc($\cdot$) and the expressions of $A_{1,2,3}(K_1, K_2)$ are very complicated. This is difficult for the application of Kriging-based methods according to the experiment study. In the implementation of RU-SAIS, the number of Gaussian densities is set as $K=15$, and the numbers of samples from the intermediate Gaussian mixture densities and the final one are taken $N_G=750$ and $N_{1,2}=1500$, respectively. In a single run of the proposed method, estimation of $\hat{I}_1$ and $\hat{I}_2$ require both 6 intermediate densities, and respectively 3 and 2 runs of the cross entropy-based iteration algorithm. The simulation results are summarized in Table 6.

**Table 6.** Comparison of the results of subset simulation, RU-SAIS and crude MCS for example 3

| Method | Result of a single run | Number of samples | COV $(\hat{\delta}_1, \hat{\delta}_2)$ |
|---|---|---|---|
| Crude MCS | $\dfrac{2.99\cdot10^{-7}}{2.44\cdot10^{-5}} = 1.23\cdot10^{-2}$ | $1.60\cdot10^9 + 1.60\cdot10^7 = 1.62\cdot10^9$ | 4.6%, 4.6% |
| Subset simulation | $\dfrac{2.49\cdot10^{-7}}{2.60\cdot10^{-5}} = 9.58\cdot10^{-3}$ | $1.94\cdot10^5 + 9.94\cdot10^4 = 2.93\cdot10^5$ | 9.2%, 7.5% |
| RU-SAIS | $\dfrac{2.71\cdot10^{-7}}{2.39\cdot10^{-5}} = 1.13\cdot10^{-2}$ | $750\cdot11 + 1500\cdot4 + 1500\cdot3 = 1.88\cdot10^4$ | 3.4%, 3.0% |

In this example, it is shown that both $I_1$ and $I_2$ are very small and the resulting posterior failure probability is relatively large. Compared with subset simulation, the proposed RU-SAIS method achieves more accurate estimation and lower COVs with lower computational cost. It implies that the final ISD using Gaussian mixture can well approximate the posterior density enclosed by highly nonlinear limit state surfaces. Moreover, through the observation of the results of the above three examples, the computation performance of RU-SAIS is not sensitive to the orders of magnitude of $I_1$ and $I_2$ and the nonlinearity of the limit state, but will be slightly affected by the dimension of the problem. This is because $K$, $N_G$ and $N_{1,2}$ increases as the dimension increases. However, considering that importance sampling naturally does not apply to problems of too high dimensions. When the number of variables is too large, the Gaussian mixture density may not be a good approximation of the optimal ISD, and there are even problems where some covariance matrixes are close to singular, which makes the algorithm cannot converge. Therefore, several key parameters should be selected out through sensitivity analysis before using this method. As for the number of measurements, it does not directly influence the performance of RU-SAIS, but may influence the difference between the prior density and the target optimal ISD. This will slightly increase the value of $M_{1,2}$.

## 6. Conclusions

This paper proposes a robust method to tackle reliability updating problems with equality information called RU-SAIS. The method decomposes the solution of the posterior failure probability into the ratio of two integrals, which are both estimated using importance sampling. RU-SAIS uses Gaussian mixture density to construct the ISD resembling the optimal one through two key steps. These include a sequential adaptive importance sampling procedure combining the elements of sequential importance sampling and $K$-means clustering, and a cross entropy-based iteration algorithm. The RU-SAIS method is terminated when the estimated COV of the estimator is reduced below a threshold value. Three examples are investigated in depth to demonstrate the performance of RU-SAIS, which shows that the results of the proposed method can achieve the same accuracy as crude MCS with much lower computational cost. Compared with subset simulation, RU-SAIS enables to obtain more robust estimators when the computational cost is in the same or even lower orders of magnitude and it is more suitable for dynamic updating problems where information is available in sequence. Moreover, another advantage of the RU-SAIS is that the computation performance is almost not affected by the number of measurements, nonlinearity in the limit state function and the level of failure probability. The main limitation is that the proposed method is based on AIS and thus naturally does not apply to very high dimensional problems. As further improvement, other typical parametric densities such as von Mises-Fisher mixture can be integrated with this method to improve its computational performance in high dimensional problems.

## Acknowledgements

This research was supported in part the National Key Research and Development

Program of China (2019YFE0112800), and by a grant (2021GQC0003) from the Institute for Guo Qiang, Tsinghua University. These supports are greatly appreciated.

## Appendix 1

Assume that $N_G$ samples from $p_G^{(i-1)}(x_g)$, denoted as $\{x_g^{(k)}: k=1,2,...,N_G\}$, are available, one can compute the weights of these samples with respect to $p^{(i-1)}(x_g)$, i.e.,

$$\omega_1(x_g^{(k)}) = \frac{p^{(i-1)}(x_g^{(k)})}{p_G^{(i-1)}(x_g^{(k)})} = \frac{\eta^{(i-1)}(x_g^{(k)})/P^{(i-1)}}{p_G^{(i-1)}(x_g^{(k)})}. \tag{1-1}$$

Term $p^{(i)}(x_g)/p^{(i-1)}(x_g)$ as the relative weight and denote it as $\omega_r(x_g)$, i.e.,

$$\omega_r(x_g) = \frac{p^{(i)}(x_g)}{p^{(i-1)}(x_g)} = \frac{\eta^{(i)}(x_g)/P^{(i)}}{\eta^{(i-1)}(x_g)/P^{(i-1)}}. \tag{1-2}$$

The estimated mean and standard deviation of $\omega_r(x_g)$ in terms of the samples can be derived as

$$\begin{aligned}
\hat{\mu}_{\omega_r} &= \frac{\sum_{k=1}^{N_G} \omega_r(x_g^{(k)}) \omega_1(x_g^{(k)})}{\sum_{k=1}^{N_G} \omega_1(x_g^{(k)})} \\
&= \frac{\sum_{k=1}^{N_G} p^{(i)}(x_g^{(k)})/p_G^{(i-1)}(x_g^{(k)})}{\sum_{k=1}^{N_G} p^{(i-1)}(x_g^{(k)})/p_G^{(i-1)}(x_g^{(k)})} \\
&= \frac{P^{(i-1)}}{P^{(i)}} \frac{\sum_{k=1}^{N_G} \eta^{(i)}(x_g^{(k)})/p_G^{(i-1)}(x_g^{(k)})}{\sum_{k=1}^{N_G} \eta^{(i-1)}(x_g^{(k)})/p_G^{(i-1)}(x_g^{(k)})}
\end{aligned} \tag{1-3}$$

and

$$\hat{\sigma}_{\omega_r} = \sqrt{\frac{\sum_{k=1}^{N_G}\left[\omega_r\left(\bm{x}_g^{(k)}\right)-\hat{\mu}_{\omega_r}\right]^2 \omega_1\left(\bm{x}_g^{(k)}\right)}{\sum_{k=1}^{N_G}\omega_1\left(\bm{x}_g^{(k)}\right)}}$$

$$= \sqrt{\frac{\sum_{k=1}^{N_G}\left\{\frac{P^{(i-1)}}{P^{(i)}}\left[\frac{\eta^{(i)}\left(\bm{x}_g^{(k)}\right)}{\eta^{(i-1)}\left(\bm{x}_g^{(k)}\right)}-\frac{\sum_{k=1}^{N_G}\eta^{(i)}\left(\bm{x}_g^{(k)}\right)/p_G^{(i-1)}\left(\bm{x}_g^{(k)}\right)}{\sum_{k=1}^{N_G}\eta^{(i-1)}\left(\bm{x}_g^{(k)}\right)/p_G^{(i-1)}\left(\bm{x}_g^{(k)}\right)}\right]\right\}^2 \frac{\eta^{(i-1)}\left(\bm{x}_g^{(k)}\right)/P^{(i-1)}}{p_G^{(i-1)}\left(\bm{x}_g^{(k)}\right)}}{\sum_{k=1}^{N_G}\frac{\eta^{(i-1)}\left(\bm{x}_g^{(k)}\right)/P^{(i-1)}}{p_G^{(i-1)}\left(\bm{x}_g^{(k)}\right)}}}$$

$$= \frac{P^{(i-1)}}{P^{(i)}}\sqrt{\frac{\sum_{k=1}^{N_G}\left\{\left[\frac{\eta^{(i)}\left(\bm{x}_g^{(k)}\right)}{\eta^{(i-1)}\left(\bm{x}_g^{(k)}\right)}-\frac{\sum_{k=1}^{N_G}\eta^{(i)}\left(\bm{x}_g^{(k)}\right)/p_G^{(i-1)}\left(\bm{x}_g^{(k)}\right)}{\sum_{k=1}^{N_G}\eta^{(i-1)}\left(\bm{x}_g^{(k)}\right)/p_G^{(i-1)}\left(\bm{x}_g^{(k)}\right)}\right]\right\}^2 \frac{\eta^{(i-1)}\left(\bm{x}_g^{(k)}\right)}{p_G^{(i-1)}\left(\bm{x}_g^{(k)}\right)}}{\sum_{k=1}^{N_G}\eta^{(i-1)}\left(\bm{x}_g^{(k)}\right)/p_G^{(i-1)}\left(\bm{x}_g^{(k)}\right)}},$$

(1-4)

respectively. The estimated COV of the relative weight is obtained by dividing Eq. (1-4) by Eq. (1-3), and the common coefficient, $P^{(i-1)}/P^{(i)}$, is eliminated.

## Appendix 2

Consider a random variable $T_1 = (aU_1 + 1)/(bU_2 + 1)$, where $U_1$, $U_2$ are independent standard normal variables, and $a$, $b$ are positive constants. Denote $U_2$ as $T_2$, and the inverse function of $\begin{cases} t_1 = (au_1+1)/(bu_2+1) \\ t_2 = u_2 \end{cases}$ is $\begin{cases} u_1 = \left[(bt_2+1)t_1 - 1\right]/a \\ u_2 = t_2 \end{cases}$, whose Jacobian determinant is $J = \begin{vmatrix} (bt_2+1)/a & bt_1/a \\ 0 & 1 \end{vmatrix} = (bt_2+1)/a$. Therefore, the joint PDF of $(T_1, T_2)$ is derived as

$$p(t_1, t_2) = \varphi\left\{\left[(bt_2+1)t_1 - 1\right]/a\right\}\varphi(t_2)|J|$$
$$= \frac{1}{2\pi a}\exp\left[-\frac{\left\{\left[(bt_2+1)t_1 - 1\right]/a\right\}^2 + t_2^2}{2}\right]|bt_2+1|.$$

(2-1)

The PDF of $T_1$ is thus

$$p_{T_1}(t_1) = \int_{t_2} p(t_1, t_2) dt_2$$

$$= \int_{t_2} \frac{1}{2\pi a} \exp\left[-\frac{\{[(bt_2+1)t_1 - 1]/a\}^2 + t_2^2}{2}\right] |bt_2 + 1| dt_2$$

$$= \frac{1}{2\pi a} \int_{t_2} \exp\left[-\frac{(b^2 t_1^2 + a^2) t_2^2 + 2bt_1(t_1-1)t_2 + (t_1-1)^2}{2a^2}\right] |bt_2 + 1| dt_2$$

$$= \frac{1}{2\pi a} \int_{t_2} \exp\left[-\frac{(b^2 t_1^2 + a^2)\left[t_2 + \frac{bt_1(t_1-1)}{b^2 t_1^2 + a^2}\right]^2 + \frac{a^2(t_1-1)^2}{b^2 t_1^2 + a^2}}{2a^2}\right] |bt_2 + 1| dt_2$$

$$= \frac{1}{2\pi a} \exp\left[-\frac{(t_1-1)^2}{2(b^2 t_1^2 + a^2)}\right] \int_{t_2} \exp\left[-\frac{(b^2 t_1^2 + a^2)\left[t_2 + \frac{bt_1(t_1-1)}{b^2 t_1^2 + a^2}\right]^2}{2a^2}\right] |bt_2 + 1| dt_2. \quad (2\text{-}2)$$

Set $t_2' = t_2 + \frac{bt_1(t_1-1)}{b^2 t_1^2 + a^2}$, and then

$$p_{T_1}(t_1) = \frac{1}{2\pi a} \exp\left[-\frac{(t_1-1)^2}{2(b^2 t_1^2 + a^2)}\right] \int_{t_2'} \exp\left[-\frac{(b^2 t_1^2 + a^2) t_2'^2}{2a^2}\right] \left|b\left[t_2' - \frac{bt_1(t_1-1)}{b^2 t_1^2 + a^2}\right] + 1\right| dt_2'$$

$$= \frac{1}{2\pi a} \exp\left[-\frac{(t_1-1)^2}{2(b^2 t_1^2 + a^2)}\right] \int_{t_2'} \exp\left[-\frac{(b^2 t_1^2 + a^2) t_2'^2}{2a^2}\right] \left|bt_2' + \frac{b^2 t_1 + a^2}{b^2 t_1^2 + a^2}\right| dt_2'. \quad (2\text{-}3)$$

Set $a_t = \sqrt{t_1^2/a^2 + 1/b^2}$, $b_t = t_1/a^2 + 1/b^2$, $c = 1/a^2 + 1/b^2$,

$d_t = \exp\left[-\frac{(t_1-1)^2}{2(b^2 t_1^2 + a^2)}\right] = \exp\left(\frac{b_t^2 - c a_t^2}{2a_t^2}\right)$, thus

$$\begin{aligned}
p_{T_1}(t_1) &= \frac{d_t}{2\pi a}\int_{t_2'}\exp\left(-\frac{b^2 a_t^2 t_2'^2}{2}\right)\left|bt_2' + \frac{b_t}{a_t^2}\right|dt_2' \\
&= \frac{d_t}{2\pi ab}\left[\int_{bt_2'=-b_t/a_t^2}^{bt_2'=\infty}\exp\left(-\frac{a_t^2(bt_2')^2}{2}\right)\left(bt_2' + \frac{b_t}{a_t^2}\right)d(bt_2') \right. \\
&\quad \left. -\int_{bt_2'=-\infty}^{bt_2'=-b_t/a_t^2}\exp\left(-\frac{a_t^2(bt_2')^2}{2}\right)\left(bt_2' + \frac{b_t}{a_t^2}\right)d(bt_2')\right] \\
&= \frac{d_t}{2\pi ab}\left\{\int_{bt_2'=-b_t/a_t^2}^{bt_2'=\infty}\exp\left(-\frac{a_t^2(bt_2')^2}{2}\right)d\left(\frac{(bt_2')^2}{2}\right) - \int_{bt_2'=-\infty}^{bt_2'=-b_t/a_t^2}\exp\left(-\frac{a_t^2(bt_2')^2}{2}\right)d\left(\frac{(bt_2')^2}{2}\right) \right. \\
&\quad \left. + \frac{b_t}{a_t^3}\left[\int_{-b_t/a_t}^{\infty}\exp\left(-\frac{(ba_t t_2')^2}{2}\right)d(ba_t t_2') - \int_{-\infty}^{-b_t/a_t}\exp\left(-\frac{(ba_t t_2')^2}{2}\right)d(ba_t t_2')\right]\right\} \\
&= \frac{d_t}{2\pi ab}\left\{-\frac{1}{a_t^2}\left[\left.\exp\left(-\frac{a_t^2(bt_2')^2}{2}\right)\right|_{bt_2'=-b_t/a_t^2}^{bt'=\infty} - \left.\exp\left(-\frac{a_t^2(bt_2')^2}{2}\right)\right|_{bt_2'=-\infty}^{bt_2'=-b_t/a_t^2}\right] \right. \\
&\quad \left. + \frac{\sqrt{2\pi}b_t}{a_t^3}\left[\Phi(b_t/a_t) - \Phi(-b_t/a_t)\right]\right\} \\
&= \frac{d_t}{a_t^2 \pi ab}\exp\left(-b_t^2/2a_t^2\right) + \frac{b_t d_t}{a_t^3\sqrt{2\pi}ab}\left[\Phi(b_t/a_t) - \Phi(-b_t/a_t)\right] \\
&= \frac{1}{a_t^2 \pi ab}\exp(-c/2) + \frac{b_t d_t}{a_t^3\sqrt{2\pi}ab}\left[\Phi(b_t/a_t) - \Phi(-b_t/a_t)\right].
\end{aligned} \tag{2-4}$$